%% file: IEEECOINS_format/main_ieee.tex
\documentclass[conference]{IEEEtran}
\IEEEoverridecommandlockouts
\usepackage{cite}
\usepackage{amsmath,amssymb,amsfonts}
\usepackage{algorithmic}
\usepackage{graphicx}
\usepackage{textcomp}
\usepackage{xcolor}
\renewcommand\IEEEkeywordsname{Keywords}
\usepackage[utf8]{inputenc} 
\usepackage[T1]{fontenc}    
\usepackage{hyperref}       
\usepackage{url}            
\usepackage{booktabs}       
\usepackage{amsfonts}       
\usepackage{nicefrac}       
\usepackage{microtype}      
\usepackage{lipsum}
\usepackage{graphicx}
\usepackage{multicol,multirow}
\usepackage{color}
\usepackage{subcaption}
\usepackage{amsmath}
\newtheorem{proposition}{Proposition}

\usepackage{algorithm}

\usepackage{algorithmic}
\usepackage{graphicx}
\graphicspath{{../figures}}     
\usepackage{multicol,multirow}
\usepackage{color}
\usepackage{subcaption}
\usepackage{newfloat}
\usepackage{listings}

\def\BibTeX{{\rm B\kern-.05em{\sc i\kern-.025em b}\kern-.08em
    T\kern-.1667em\lower.7ex\hbox{E}\kern-.125emX}}
\begin{document}

\title{PropEnc: A Property Encoder for Graph Neural Networks
}
\author{\IEEEauthorblockN{Anwar Said}
\IEEEauthorblockA{\textit{Department of Computer Science} \\
\textit{Vanderbilt University}\\
Nashville, TN, USA \\
anwar.said@vanderbilt.edu}
\and
\IEEEauthorblockN{Waseem Abbas}
\IEEEauthorblockA{\textit{Systems Engineering Department} \\
\textit{University of Texas at Dallas}\\
Richardson, TX, USA \\
waseem.abbas@utdallas.edu}
\and
\IEEEauthorblockN{Xenofon Koutsoukos}
\IEEEauthorblockA{\textit{Department of Computer Science} \\
\textit{Vanderbilt University}\\
Nashville, TN, USA \\
xenofon.koutsoukos@vanderbilt.edu}
\thanks{This material is based upon work supported by the National Science
Foundation under Grant Nos. 2325416 and 2325417.}
}

\maketitle

\IEEEpubid{\makebox[\columnwidth]{\copyright~979-8-3315-2037-3/25/\$31.00 ©2025 IEEE. } \hspace{\columnsep}\makebox[\columnwidth]{ }}

\begin{abstract}

Graph machine learning, particularly using graph neural networks, heavily relies on node features. However, many real-world systems, such as social and biological networks, lack node features due to privacy concerns, incomplete data, or collection limitations. Structural and positional encoding are commonly used to address this but are constrained by the maximum values of the encoded properties, such as the highest node degree. This limitation makes them impractical for scale-free networks and applications involving large or non-categorical properties. This paper introduces \emph{PropEnc}, a novel and versatile encoder to generate expressive node embedding from any graph metric. By combining histogram construction with reversed index encoding, \emph{PropEnc} offers a flexible solution that supports low-dimensional representations and diverse input types, effectively mitigating sparsity issues while improving computational efficiency. Additionally, it replicates one-hot encoding or approximates indices with high accuracy, making it adaptable to a wide range of graph applications. We validate \emph{PropEnc} through extensive experiments on graph classification task across several social networks lacking node features. The empirical results demonstrate that \emph{PropEnc} offers an efficient mechanism for constructing node features from various graph metrics.
\end{abstract}

\IEEEpubidadjcol
\input{sections}
\bibliographystyle{ieeetr}  
\bibliography{references}

\end{document}

%% file: sections.tex
\section{Introduction}
\label{sec:introduction}

Graph Neural Networks (GNNs) rely heavily on node features due to their inherent message passing mechanisms, which necessitate rich node representations \cite{kipf2016semi}. In GNNs, the message passing process involves the iterative exchange of information between neighboring nodes. 
This iterative process ensures that the node embeddings capture the structural and feature-based information from their local neighborhoods. The expressiveness of the initial node features is thus crucial; more informative node features lead to more meaningful node embeddings, ultimately enhancing the performance of the GNN on various tasks, such as node classification, link prediction, and graph classification \cite{hamilton2017inductive,said2023enhanced}. However, many real-world systems, including social networks, financial networks, and communication networks often lack node features due to various challenges \cite{duong2019node}. For instance, in social networks, data may be incomplete or missing; users may not have provided certain personal details or activities. In financial networks, missing data issues may arise due to transactional privacy or unreported financial activities. Similarly, in communication networks, device-level data such as signal strength or connection history might be unavailable due to privacy settings or technical constraints.

In featureless networks, researchers typically resort to positional and structural node features to compensate for the lack of inherent features \cite{cui2022positional}. Positional features encode the nodes' positions (location relative to other nodes) within the graph, with examples including random features, eigen decomposition, and deep walk features \cite{duong2019node}. Conversely, structural features capture the topological properties of the nodes, with degree and PageRank being such examples. In various settings, structural features such as degree are usually encoded using one-hot encoding \cite{xu2018powerful}. Similarly, other features are typically stacked into feature vectors \cite{parkes2021network}. This approach, however, poses several challenges, most notably the issue of \emph{high dimensionality}. For instance, when encoding the degree as features, the dimensionality of the feature vector depends on the maximum degree within the dataset, which can be in the thousands. This leads to \emph{sparsity of the embeddings}. Additionally, in settings such as PageRank, Egonet size, etc., the feature for each node would be a \emph{single integer or decimal value}. Having just one value as a node feature may not enhance model performance \cite{cui2022positional}. If the corresponding value is decimal, such as the PageRank and betweenness centrality among many others, \emph{one-hot encoding might not be a suitable solution} for that. These limitations necessitate the development of more efficient and expressive encoding schemes to improve the applicability and performance of the machine learning (ML) models.

So a natural question arises: \textit{how can we encode any arbitrary graph metric that remains independent of both the predefined dimensional size and the nature of the properties, whether structural or positional, categorical or decimal value?} In this work, we propose \textit{Property Encoder} (\textit{PropEnc}), a versatile method for transforming any type of graph metric into node features. Unlike traditional methods that require a fixed feature size, \textit{PropEnc} treats this size as a hyper-parameter, thereby offering greater flexibility. Moreover, it can process properties of any type, including structural or positional metrics, as well as categorical or continuous values. Our results indicate that encoding large social networks with PropEnc using dimensions of $50$ or even fewer not only achieves superior or comparable performance but also significantly reduces the dimensionality of embeddings—from potentially thousands (corresponding to the maximum node degree) to just a few dimensions. This reduction substantially decreases the number of parameters, enhancing the efficiency and scalability of the model. Additionally, by encoding several other graph contrality measures, we show that PropEnc can be a mean to encode other metrics that may lead to improved performance in different settings. The core idea of \textit{PropEnc} lies in its use of histogram representation to encode any given graph metric. Specifically, it leverages reverse indexing to determine the corresponding indices for each node based on the range in which their values fall within the histogram. This novel scheme enables the encoding of any kind of graph metric while preserving positional information at both local and global graph levels. The main contributions of this work are as follow:

\begin{itemize}
    \item We introduce \textit{PropEnc}, a universal encoding scheme designed to handle any type of graph metrics. Our analysis demonstrates that this scheme can generate node features that either exactly replicate or closely approximate indices across various settings.
    \item We perform a series of experiments to demonstrate that \textit{PropEnc} is an effective method for graph machine learning, particularly on networks lacking inherent node features.
\end{itemize}

\textit{PropEnc} can be employed in a wide range of applications, including social networks, bioinformatics, and financial network, where node features are often missing or incomplete. Its ability to handle diverse metrics without requiring predefined fixed features size makes it particularly useful in scenarios such as large-scale graphs and scale-free networks where structural encoding (degree) leads to large dimensional node features. Furthermore, PropEnc offers a potentially superior method for aligning node features to optimize the training of graph foundation models. This is particularly advantageous given that these models are trained on diverse datasets spanning various fields.

\section{Related Work}
\label{sec:related-work}

Several features initialization methods have been studied in literature to generate expressive node features in unattributed networks. Broadly, these approaches can be categorized into two categories: centrality-based approaches and learning-based approaches \cite{cui2022positional}. In the forthcoming sections, we provide a brief overview of these approaches. 

\subsection{Centrality-based approaches}

Centrality-based approaches predominantly focus on leveraging node roles or structural properties to construct feature sets. In featureless networks, various centrality-based metrics have been used that generate informative scalar quantities. Examples of these metrics include degree centrality, Egonet size, the number of triangles, k-core number, PageRank, betweenness centrality, and closeness centrality \cite{rossi2017deep,xu2018powerful,henderson2012rolx,said2024improving,chen2019exploiting}. Among these, degree centrality is a prevalent method for feature initialization, where each node's feature is represented by a one-hot encoded vector corresponding to its degree \cite{xu2018powerful}. The encoding process begins by identifying the maximum degree within the dataset, followed by constructing the one-hot encoded vectors based on the degree of each node. These encoded vectors are subsequently used as node features and input into GNNs for networks lacking inherent features \cite{xu2018powerful,hamilton2017inductive}. Beyond degree centrality, other measures such as PageRank, betweenness centrality, and closeness centrality offer valuable insights into node significance and graph structure and have been used in recent studies \cite{parkes2021network, chen2019exploiting,said2024improving}. 

A further strategy is the construction of multi-centrality feature vectors, combining various centrality scores into a singular feature vector for each node \cite{parkes2021network,said2024improving}. This combined approach aims to exploit the complementary strengths of different centrality metrics, yielding a holistic representation of node importance \cite{henderson2012rolx,chen2019exploiting}. These methodologies are primarily categorized under structural properties. In addition, there have been efforts to incorporate positional encoding schemes, such as the generation of random features \cite{errica2019fair,you2019position} using distributions determined by random seeds, to represent nodes in a high-dimensional space. Although these random encoding do not directly reflect the relative positions of the nodes, they can aid GNNs in implicitly learning node positions \cite{cui2022positional}. 

\subsection{Learning-based approaches}

In learning-based approaches, node features are often derived as node embeddings obtained through unsupervised learning processes that take into account the entire graph structure. For example, methods like DeepWalk \cite{perozzi2014deepwalk}, Node2Vec \cite{grover2016node2vec} and HOPE \cite{ou2016asymmetric} represent prominent unsupervised embedding techniques. DeepWalk and Node2Vec are shallow embedding techniques where two nodes are deemed to be close if they frequently co-occur on random walks originating from either node. On the other hand, HOPE can be viewed as a graph factorization method, where variants of the adjacency matrix are factorized to produce node embeddings. Additionally, eigendecompositions serve as another technique to capture the spectral properties of graphs, thereby facilitating the embedding of nodes into lower-dimensional spaces. These approaches collectively enable the generation of meaningful representations of nodes that encapsulate intricate structural details and relationships within the graph \cite{perozzi2014deepwalk,ou2016asymmetric}.\cite{chaudhuri2012spectral,perozzi2014deepwalk}. 


These methodologies offer several strategies for constructing node features. However, they encounter limitations when vector representations are needed instead of scalar values. While one-hot encoding is a straightforward approach, it becomes impractical when the range of values is extensive, as it relies on the maximum value within the dataset. Additionally, one-hot encoding struggles with representing metrics that yield decimal numbers. To address these challenges, we propose a novel encoder function designed to overcome these limitations effectively. 

\section{Preliminaries}
In networks lacking inherent node features, various node properties and their corresponding embeddings have been leveraged to initialize node feature representations \cite{cui2022positional}. More informative node features can provide a stronger starting point for GNNs, thereby enhancing performance on subsequent downstream tasks. In the following sections, we first introduce the reader with relevant and quite well-known concepts and then we present the proposed encoder that can be used to expressively encode any type of node property.

\subsection{Features Encoding Schemes}
\label{subsec:features-encoding-schemes}

\textbf{One-hot encoding:} Feature encoding schemes play a vital role in data preprocessing for machine learning models. Among these, one-hot encoding is particularly prevalent due to its simplicity and effectiveness in managing categorical data. One-hot encoding translates categorical variables into binary vectors, facilitating easier interpretation by machine learning models.

\begin{figure}[!htb]
    \centering
    \includegraphics[width=0.95\linewidth]{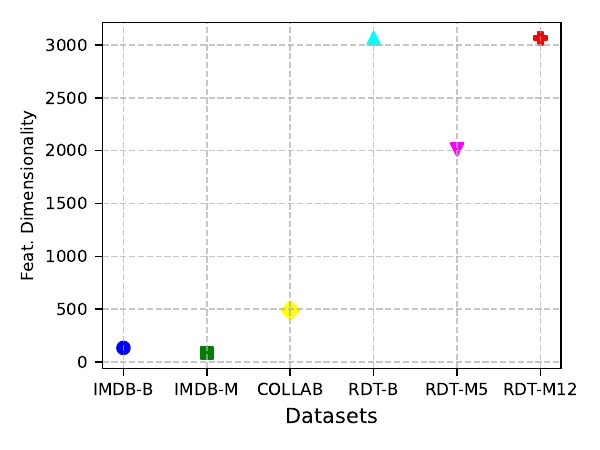}
    \vspace{-0.1in}
    \caption{Illustration of features dimensionality for five well-known featureless graph classification datasets. For large datasets such as Reddit (RDT), where the graph sizes reach thousands of nodes, the feature dimensionality correspondingly escalates significantly. Specifically, the maximum number of nodes (largest graph) in these datasets are $136,89,492,3782,3648,3782$, respectively. }
    \label{fig:feat_dim}\vspace{-0.2in}
\end{figure}

One-hot encoding is formally described as follows: Given a categorical variable with \(k\) distinct categories, the \(i\)-th category is represented by a \(k\)-dimensional vector \(\mathbf{v}_i\), defined as:
\[
\mathbf{v}_i = \begin{cases} 
    1 & \text{if } j = i, \\ 
    0 & \text{otherwise,} 
    \end{cases}
\]
for \( j = 1, 2, \ldots, k \). This transformation ensures a sparse representation that prevents any spurious ordinal relationships between categories, thereby preserving their distinctiveness. In graph representation learning, we usually encode graph metrics such as degree, where the one-hot encoding is just a $d-$dimensional degree encoded feature vector. $d$ corresponds to the maximum degree in the entire dataset. One-hot encoding offers several advantages, including the conversion of categorical data into a numerical format that is more readily processed by machine learning models. Additionally, it effectively avoids the problem of implying ordinality among categories, potentially enhancing model performance by providing a more accurate representation of the input data. 
However, when dealing with scale-free networks, one-hot encoding can become highly problematic. For instance, in degree encoding scenario, a small number of nodes have very high degrees (such as influential users), while the vast majority have relatively low degrees. This results in a highly skewed distribution with a large range of values. Since one-hot encoding translates each unique degree value into a separate binary vector, with the length of the vector equal to the maximum degree in the graph, this leads to an explosion in the dimensionality of the encoded feature space. In Figure \ref{fig:feat_dim}, we present the feature dimensional space for 5  well-known featureless graph classification datasets \cite{Morris+2020} to highlight the feature dimensional space when using one-hot degree encoding. We can see that the feature space significantly increases when the size of the graphs in the datasets increases.  

 The high dimensionality intrinsic to one-hot encoding not only makes the encoding computationally and memory expensive but also worsens issues of data sparsity. This is because most nodes will have a degree represented by a '1' in a single position, and '0's elsewhere. For instance, in the Reddit-Binary dataset where the maximum degree is 3062, each node will have a feature vector of size 3062, with only one index as '1' and all others as '0's. This sparsity can hinder the learning process in downstream machine learning tasks, as evidenced by our empirical evaluation and several other studies.

Another major challenge of one-hot encoding is its inability to represent decimal values, as it is designed exclusively for categorical data. We believe this limitation may explain why researchers in the field often rely solely on degree to construct node features in featureless networks. However, there are numerous network science metrics that could be leveraged to construct node features, each offering a unique perspective. Examples include PageRank, centrality metrics such as betweenness, closeness, and many others. Therefore, there is a pressing need for a method that not only considers various metrics but is also flexible in terms of dimensionality and expressive enough to maintain the accuracy of the models.


\textbf{Histogram Representation:} A histogram is a powerful technique to capture the frequency distribution of any given data modality or certain attributes. It is a graphical representation that organizes a group of data points into user-specified ranges (bins) and allows a flexible way to avoid the high dimensionality issue. By aggregating individual data points into bins, histograms succinctly capture the overall distribution of the data, highlighting key patterns and trends without losing critical information \cite{ruppert2004elements}.

This ability to condense and summarize data makes histograms particularly useful in graph representation learning, which leverage these distributions to construct graph representations. For instance, the Weisfeiler-Lehman Subtree Kernel \cite{shervashidze2011weisfeiler} utilizes a histogram-based representation to count subtree patterns across different graphs, facilitating the measurement of graph similarity. Similarly, several other studies use histograms to construct final graph representations for the downstream ML task such as graph classification \cite{verma2017hunt,borgwardt2005shortest}. By reflecting the global distribution of structural features, histograms enable graph kernels to represent and differentiate graphs in a way that is both computationally efficient and expressive, ultimately enhancing the performance and interpretability of graph machine learning models. 

Leveraging the potential of histograms, we can effectively construct graph-level representations that capture the distribution of various properties within the representation space. For example, histograms can be employed to represent shortest path distances, counts of subtrees, or graphlets. However, when focusing on node-level features derived from graph metrics such as standard degree or PageRank, histograms fall short, necessitating the use of traditional one-hot encodings. These encodings often suffer from issues related to high dimensionality and sparsity. In the following, we propose a novel scheme that utilizes histograms for constructing node features.

\section{PropEnc: A Versatile Encoding Scheme}
\label{sec:method}

Let $G = (V, E)$ be a graph, where $V = \{v_1, v_2, \ldots v_n\}$ is the set of nodes and $E$ is the set of edges. Let $\phi: V \rightarrow \mathcal{P}$ such that $\mathcal{P}^{|V| \times 1}$ is a set of values containing one for each node obtained through a graph metric, $\Phi$. We define histogram $h_G(\mathcal{P})$ with range $ = (min(\mathcal{P}), max(\mathcal{P}))$ with $d$ number of bins on top of $\mathcal{P}$ to construct graph-level representation $h_G$. Note that $h_G$ is the representation of the relative frequencies of the property $\mathcal{P}$ not in regards to the given graph $G$ but in regards to the entire search space (e.g., complete dataset). Due to this global-level representation, $h_G$ keeps the entire distribution intact.

Given $h_G$, we define node representation $h_v$ for a node $v$ based on reversed one-hot frequency indexing. Formally, we define $h_v$ as follows. 

\begin{equation*}
\small
\mathbf{h}_v(i) = 
\begin{cases} 
1 & \text{if} \hspace{0.05in} \phi(v) \in h_G(i) \\
0 & \text{otherwise}
\end{cases}
\end{equation*}

$h_v$ is a one-hot encoded representation of $\mathcal{P}$ which transforms the global-level graph-level representation into a condensed node-level representation. Note that $\phi(v)$ here returns the property, for instance the degree of node $v$ and $h_G(i)$ indicates the index of the bin where degree of node $v$ falls. Since the range of $\mathcal{P}$ could be very large for certain properties, this encoding provides a flexible way to obtain lower-dimensional representations while capturing approximately similar distribution. 

\begin{table*}
    \centering
    \begin{tabular}{|l|c|c|c|c|c|c|} \hline 
         Dataset&  Degree&  PropEnc (10)&  PropEnc (20)&  PropEnc (30)&  PropEnc (40)& PropEnc (50)\\ \hline 
         IMDB-BINARY& {\color{blue}$77.80$} $\pm2.8$ & 72.20 $\pm2.6$ & 72.00 $\pm4.5$&70.60 $\pm2.3$ &  73.20$\pm2.5$& 74.00 $\pm1.7$  \\ \hline 
         IMDB-MULTI& 42.13$\pm2.5$ & 40.00 $\pm2.7$ & 41.33$\pm2.5$ & {\color{blue}44.40} $\pm2.8$& 44.13 $\pm2.2$&38.13 $\pm1.5$\\ \hline 
         COLLAB& {\color{blue}76.56 }$\pm 1.6$ & 72.52 $\pm1.0$& 72.16 $\pm0.8$& 72.36 $\pm1.1$ & 72.16 $\pm1.4$ &72.88 $\pm 0.5$\\ \hline 
         REDDIT-BINARY& 82.40 $\pm3.1$ & $91.30 \pm1.5$ &$89.80 \pm0.8$ &$89.60 \pm1.6$ & $89.80 \pm2.0$ & {\color{blue}$91.60 \pm0.7$}\\ \hline 
         REDDIT-MULTI-5K& 45.84 $\pm 2.1$& 49.32 $\pm2.3$ & 48.92 $\pm2.4$& {\color{blue}51.40} $\pm1.4$& 49.56 $\pm2.5$&49.08 $\pm0.2$\\ \hline
         REDDIT-MULTI-12K&40.54  $\pm 0.7$& 44.72 $\pm1.4$ & {\color{blue}44.73} $\pm0.7$ &44.59 $\pm0.7$ & 43.57 $\pm1.7$ &44.49 $\pm1.7$ \\ \hline
    \end{tabular}
    \vspace{1ex}
    \caption{Performance comparison in terms of accuracy (with standard deviation) against the standard one-hot degree encoding. The numbers such as 10 and 20 with PropEnc represents the size of the encoding.}
    \label{tab:results-deg}
\end{table*}

\begin{table*}
\small
    \centering
    \begin{tabular}{|l|c|c|c|c|c|c|} \hline 
         Dataset&  Baseline&  PropEnc(Bet.)&  PropEnc(Clos.)& PropEnc(Eig.) &  PropEnc(PageRank) & PropEnc (Conc.) \\ \hline 
         IMDB-BINARY& 67.60 $\pm2.1$& {\color{blue}77.80} $\pm4.1$ & 74.20 $\pm3.4$& 73.80 $\pm3.7$& 75.80 $\pm3.5$&73.60 $\pm2.1$\\ \hline 
         IMDB-MULTI&  39.07 $\pm0.7$& {\color{blue}41.33} $\pm1.7$&40.93 $\pm2.1$ &41.07 $\pm2.8$  & 42.80 $\pm1.1$& 39.60 $\pm1.8$\\ \hline 
         COLLAB& 73.08 $\pm1.2$& {\color{blue}74.96} $\pm1.2$& 70.08 $\pm0.5$ & 72.72 $\pm0.1$& 73.00 $\pm1.7$ & 72.00 $\pm1.0$\\ \hline 
         REDDIT-BINARY& 91.40 $\pm1.9$& 89.50 $\pm1.4$& {\color{blue}92.10} $\pm2.1$ & 90.10 $\pm1.4$ & 89.00 $\pm1.3$&89.80 $\pm1.7$ \\ \hline 
         REDDIT-MULTI-5K& 50.72 $\pm1.7$& {\color{blue}51.24} $\pm1.8$& 49.16 $\pm2.8$ & 50.60 $\pm2.4$ & 50.76 $\pm1.2$ &47.88 $\pm1.9$\\ \hline 
         REDDIT-MULTI-12K& 45.53 $\pm1.2$& 44.74 $\pm 0.8$ & 44.31 $\pm1.8$& {\color{blue}45.82} $\pm0.7$& 44.39 $\pm1.1$&44.96 $\pm1.2$ \\ \hline
    \end{tabular}
    \caption{Performance comparison in terms of accuracy across different centrality measures. Five different bins of length 10, 20, 30, 40 and 50 were considered for encoding the measures and the best results among them are reported. The baseline considered as a feature vector with concatenated all the centrality values.}
    \label{tab:all-metrics}\vspace{-0.2in}
\end{table*}

\begin{figure}[!t]
    \centering
    \includegraphics[width=0.8\linewidth]{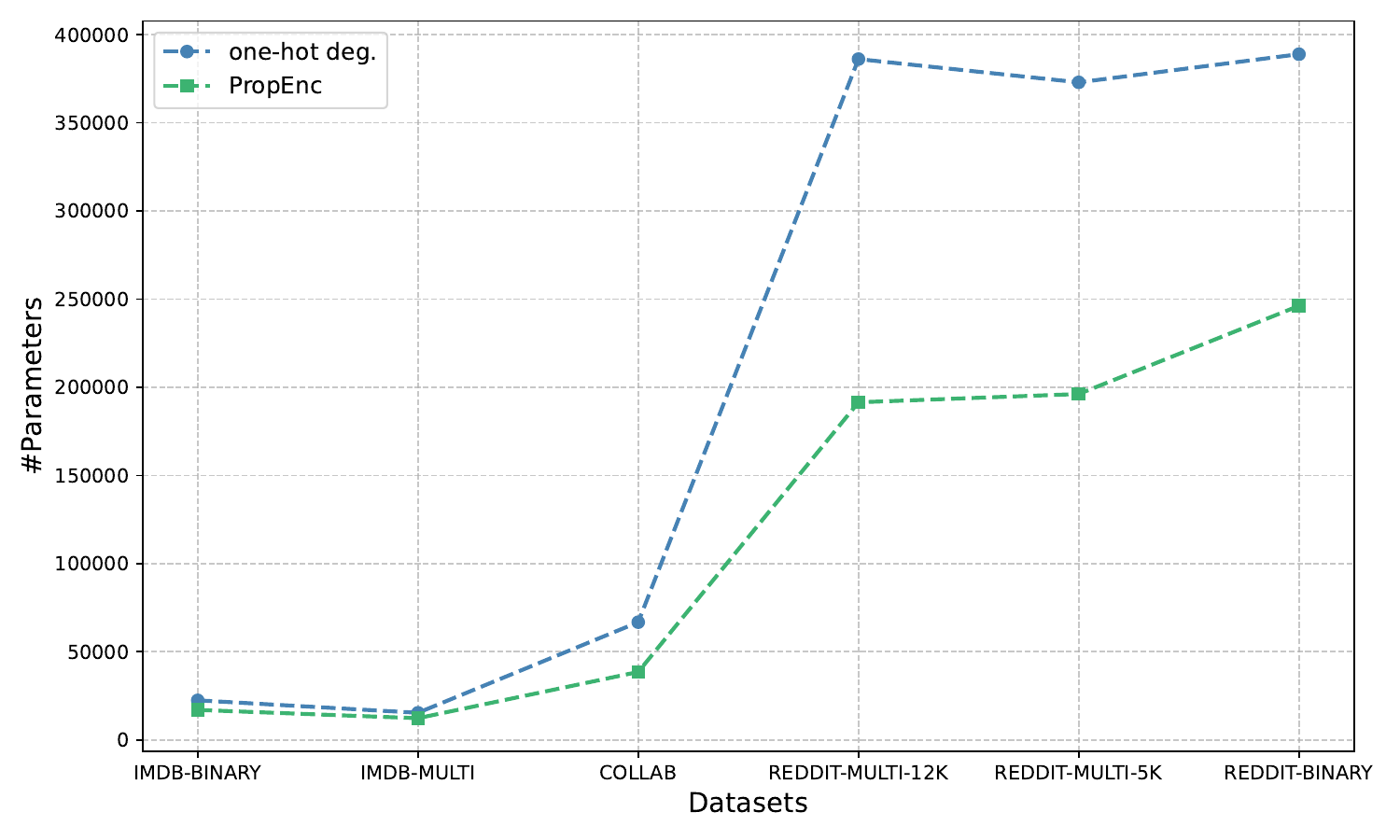}
    \caption{Comparison of the number of parameters between the proposed PropEnc and one-hot degree encoding. The increase in the dimension of features results in an increase in the number of parameters. }
    \label{fig:params-plot}\vspace{-0.2in}
\end{figure}

\begin{proposition}
\label{lemma:1}
Given a graph $G = (V, E)$, let $\phi: V \rightarrow \mathcal{P}$ be a function that assigns a property value $\mathcal{P}$ to each node $v \in V$. Let $h(\mathcal{P})$ be the histogram representation of $\mathcal{P}$ with $d$ bins and range $(1, max(\mathcal{P}))$. Then, the PropEnc's feature set $h^d_v$ is identical to the one-hot encoding of $\mathcal{P}$ if and only if $d = max(\mathcal{P})$ and the histogram range is $(1, max(\mathcal{P}))$.
\end{proposition}

This can be directly observed from the definition of one-hot encoding and the construction of the histogram representation. When the number of bins $d$ equals $max(\mathcal{P})$ and the range is $(1, max(\mathcal{P}))$, each bin in the histogram corresponds to a unique value in $\mathcal{P}$. Therefore, $h^d_v$ is identical to the one-hot encoding of the property value $\mathcal{P}(v)$ for each node $v \in V$. 

Given the aforementioned encoding scheme, PropEnc is a two-step process. In the first step, we construct the graph-level histogram representation using the specified metric, with predefined bin size and range. This approach allows us to flexibly define the size of the encoding and select the type of histogram, such as equal width, equal frequency, adaptive histograms, among others. In the second step, we generate initial node features using reverse encoding from the global graph-level histogram created in the first step. Since the global histogram aggregates nodes with similar graph metrics into the same or adjacent bins, the reverse encoding ensures that nodes with closely similar features are initialized in proximate bins.

\section{Numerical Evaluation}

To evaluate the proposed method, we consider five well-known datasets: IMDB-BINARY, IMDB-MULTI, COLLAB, REDDIT-BINARY, REDDIT-MULTI-5K, and REDDIT-MULTI-12K \cite{Morris+2020}. The reason for choosing only these datasets is that all these datasets lack node features. The first three datasets are relatively small, with maximum graph sizes of 136, 89, and 492 nodes, respectively. The Reddit datasets are considerably larger, containing 3782, 3648, and 3782 nodes in their largest graphs, respectively. Detailed statistics for all these datasets are available online and in the corresponding publication \cite{Morris+2020}. The COLLAB dataset involves a three-class classification task, while the REDDIT-MULTI-5K and REDDIT-MULTI-12K datasets pertain to five-class and eleven-class classification tasks, respectively. The remaining datasets are binary.

\textbf{Baselines:} We use degree one-hot encoding and the concatenation of four centrality metrics—betweenness centrality, closeness centrality, eigenvector centrality, and PageRank—as our baselines. Additionally, we examine the impact of different dimensionalities by varying the number of dimensions: 10, 20, 30, 40, and 50 in the proposed embeddings. For the learning model, we employ a consistent graph classification architecture to ensure a fair comparison. Our learning architecture comprises three layers of graph convolution \cite{morris2019weisfeiler}, sort pooling \cite{zhang2018end} with \( k=0.6 \), followed by two 1D convolutional layers with 1D max pooling, and finally, an MLP with three layers. The number of epochs is set to 100, learning rate to \( 1 \times 10^{-4} \), batch size to 32, and 32 hidden neurons in both the GNN convolutional and MLP layers. Dropout is set to 0.5. We use 5-fold nested cross validation with keeping 10\% final test set completely untouched throughout the experiments. We use accuracy as an evaluation metric because all the considered datasets are balanced except a small imbalance in COLLAB and REDDIT-MULTI-12K classes. We use equal width histogram throughout the experiments for constructing global-level representations during the encoding process. All experiments are conducted on a Lambda machine equipped with an AMD Ryzen Threadripper PRO 5995WX 64-Core CPU, 512 GB RAM, and an NVIDIA RTX 6000 GPU with 48 GB of memory, on Linux operating system. The implementation will be made publicly available once the paper is online. 

\subsection{Results} 

In Table \ref{tab:results-deg}, we present a comparison between the PropEnc and the standard one-hot degree encoding. The primary objective of this experiment is to evaluate the effectiveness of the encoding scheme across different dimensions on datasets containing both small and large graphs. For details on the number of one-hot encoded feature dimensions, please refer to Figure \ref{fig:feat_dim}. Our results demonstrate that the proposed encoding method significantly improves model performance, in terms of accuracy, on most datasets except for IMDB-BINARY and COLLAB. Specifically, for datasets with large graphs—which represent the primary application of this method—the performance improvements are particularly notable. Furthermore, these results indicate that the PropEnc remains effective even with relatively small encoding dimensions, such as 10 or 20 bins.

\begin{figure}[!t]
    \centering
    \includegraphics[width=0.9\linewidth]{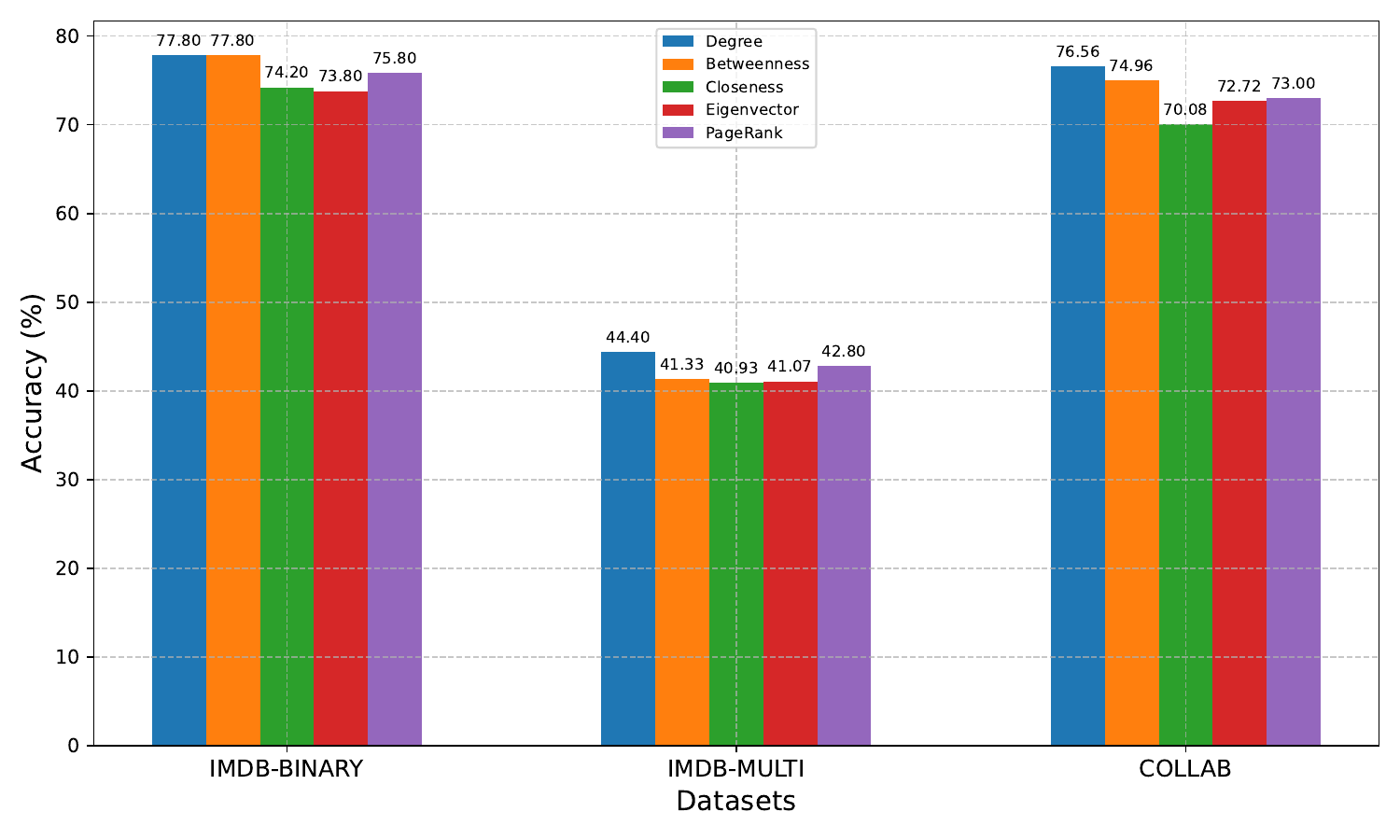}
    \caption{Performance comparison of five node metrics using a fixed GNN architecture on small, unattributed network datasets.}
    \label{fig:res-comparison1}\vspace{-0.2in}
\end{figure}

To highlight the effectiveness of compact node features, Figure \ref{fig:params-plot} presents a comparison of the number of parameters when using one-hot encoding versus the proposed PropEnc. The results clearly demonstrate that PropEnc significantly reduces the number of parameters in the model, particularly for datasets with nodes exhibiting high degrees. It is important to note that since node features serve as the input to the model, an increase in the number of input features directly increases the number of neurons in the first layer, thereby escalating the total number of parameters. Except for this variation, the rest of the architecture remains unchanged.

Furthermore, by reducing the feature dimensionality to a limited or smaller number of dimensions, PropEnc effectively decreases the sparsity of the data as well as the model's complexity as shown in Figure \ref{fig:params-plot}. This reduction in both sparsity and complexity contributes to the overall efficiency and performance of the model. Prior research has explored several measures such as the size of the egonet, k-core number, and degree, which return integer values and are thus compatible with one-hot encoding \cite{cui2022positional}. However, to the best of our knowledge, metrics that produce decimal numbers have not been systematically evaluated with GNNs.

In our subsequent experiments, we evaluate the performance of four well-known centrality measures: Betweenness, Closeness, Eigenvector, and PageRank centrality encoded with PropEnc. The reason of choosing these measures is because they are well-known and highly used for node ranking. However, one can consider any other metric or potentially new node metrics for initializing node features. We report the best results obtained among them for each dataset in Table \ref{tab:all-metrics}. Additionally, we concatenate these metrics into a single feature vector, establishing it as a comparative baseline. We also present results where binary encoded vectors of all these metrics are concatenated and then trained with GNNs. We conducted experiments for the centrality measures across five different encoding dimensions: $10, 20, 30, 40,$ and $50$. For each metric, we report the best results obtained among these dimensions.

The results of these experiments reported in Table \ref{tab:all-metrics} yield several intriguing observations. Firstly, centrality measures other than degree also demonstrate promising results, despite capturing various aspects of network information vital for GNN learning. For example, betweenness centrality, particularly shortest-path betweenness centrality, exhibits strong predictive power. Secondly, although we initially hypothesized that combining multiple metrics would enhance performance, our results indicate that concatenating embeddings from different metrics does not significantly improve model efficacy. Thirdly, eigenvector centrality and closeness centrality occasionally outperform degree encoding, illustrating that different metrics can offer unique advantages depending on the dataset and task. Additionally, our results show that when the size of the original one-hot encoding is small, such as in the case with degree encoding, it tends to perform better due to the reduced dimensional feature space and lower sparsity, which are easier for models to learn. This is corroborated by our findings on the IMDB-BINARY and COLLAB datasets. However, when the feature size expands to thousands, as observed in large networks, the performance of models using one-hot encoding diminishes significantly. In these instances, the proposed encoding scheme yields superior results.  

Moreover, our results emphasize that creating a single feature vector by concatenating multiple metrics does not contribute significantly to model performance, compared to encoding these metrics separately. This reinforces the effectiveness of PropEnc, which allows for distinct and accurate encoding of various graph metrics, thereby maintaining the model's accuracy and enhancing its performance on complex tasks. Overall, these insights underscore the versatility and effectiveness of PropEnc in broadening the spectrum of graph metrics that can be encoded and leveraged in graph-based machine learning applications, leading to improved results.

\begin{figure}[!t]
    \centering
    \includegraphics[width=0.9\linewidth]{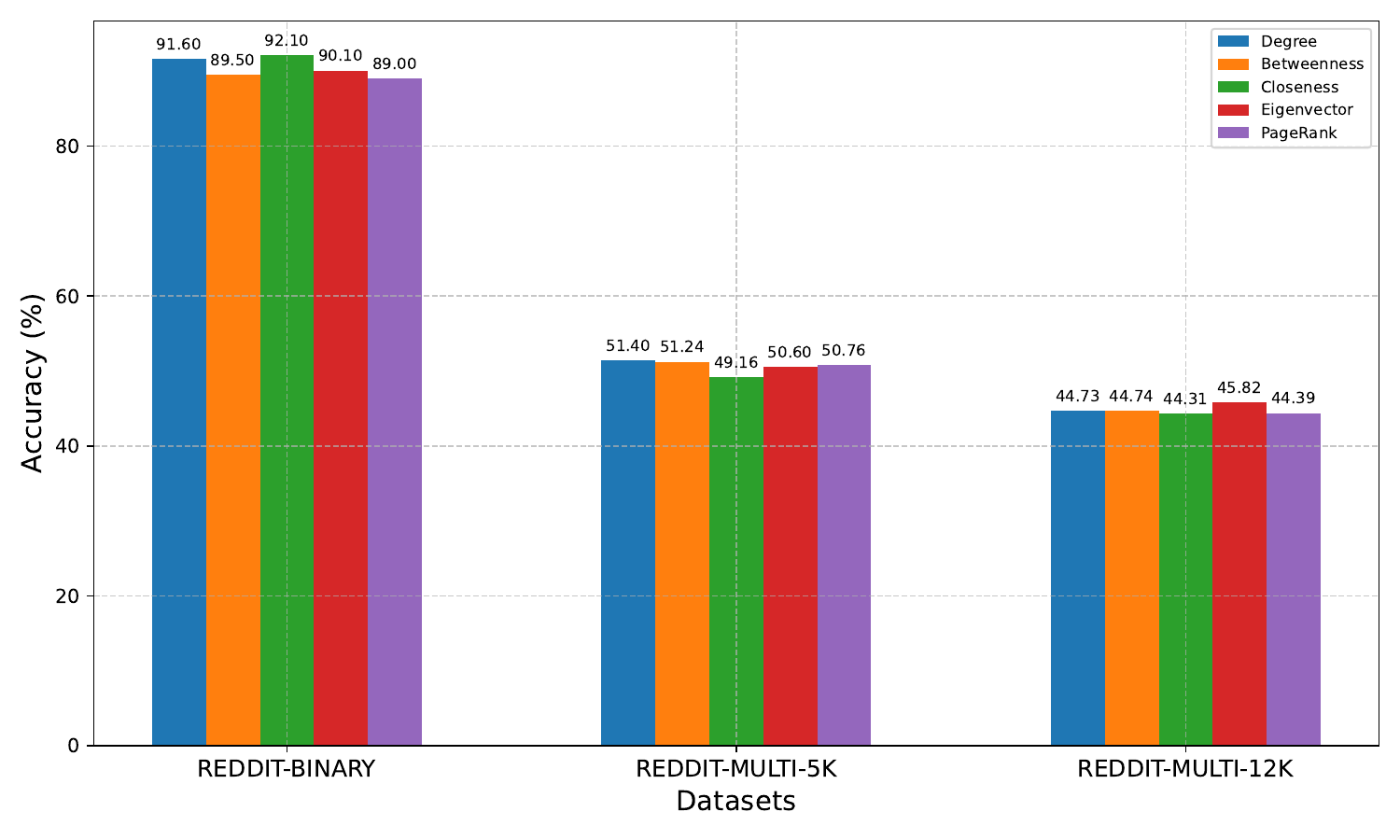}
    \caption{Performance comparison on datasets with large featureless networks. All the best results reported here were obtained through PropEnc.}
    \label{fig:res-comparison2}\vspace{-0.2in}
\end{figure}

\textbf{What is the most effective metric for initializing node features?}

We illustrate the performance comparison of five different node metrics in Figures \ref{fig:res-comparison1} and \ref{fig:res-comparison2}, highlighting the most effective feature initialization methods that researchers and practitioners may consider. The results reveal that degree still excels in many cases, achieving superior performance in 4 out of the 6 datasets. Importantly, other centrality metrics also provide sufficient information for the models to achieve competitive performance. For instance, the performance of degree and betweenness centrality is closely aligned on the IMDB-BINARY, REDDIT-MULTI-5K, and REDDIT-MULTI-12K datasets. Degree performs slightly better on the remaining three datasets. Interestingly, closeness centrality outperforms all other metrics on the REDDIT-BINARY dataset, while its performance remains lower on the others. Meanwhile, eigenvector centrality excels on the REDDIT-MULTI-12K dataset. Although the concatenation of these metrics into a single feature vector was not particularly effective, as discussed in the previous section, PropEnc enables these metrics to significantly enhance model performance. Our empirical results underscore that PropEnc, with its reduced dimensional space, markedly improves model accuracy and efficiency while also reducing memory usage.

\section{Conclusion}

Learning on featureless networks consistently necessitates effective methods for constructing features that are beneficial for downstream tasks. This area of research remains underexplored, presenting numerous opportunities for innovation. In this work, we introduce PropEnc, a universal and simple encoder that constructs expressive node features from any given graph metric. It exhibits flexibility in terms of both dimensionality and input types, including categorical, integer, and decimal values. This versatility allows for the consideration of a wide range of metrics, addressing various research questions regarding the types of metrics that contribute most effectively to the performance of GNNs. Our extensive evaluations, which compared PropEnc against standard degree encoding, centrality concatenation, and encoding concatenation with various dimensions, demonstrate the effectiveness of the proposed method. The empirical results consistently show that PropEnc improves model performance across different graph classification tasks.

Based on the proposed PropEnc, several avenues for future research can be considered. For instance, investigating which features derived from structural and positional encoding lead to improved GNN performance could provide valuable insights. Additionally, exploring the combination of encoded topological or structural information with original features to enhance performance presents another intriguing research question. 